\documentclass[10pt,twocolumn,letterpaper]{article}

\usepackage[pagenumbers]{cvpr} 
\usepackage{multirow}
\usepackage{textcomp}
\usepackage[accsupp]{axessibility}

\usepackage{algorithm}
\usepackage{algorithmic}

\definecolor{cvprblue}{rgb}{0.21,0.49,0.74}

\usepackage[pagebackref,breaklinks,colorlinks ,allcolors=cvprblue]{hyperref}

\def\paperID{1453} 
\def\confName{CVPR}
\def\confYear{2025}
\usepackage{cuted}
\title{RoomPainter: View-Integrated Diffusion for Consistent Indoor Scene Texturing}

\author{
Zhipeng Huang$^{1}$ \quad Wangbo Yu$^{1,2}$ \quad Xinhua Cheng$^{1}$ \quad Chengshu Zhao$^{1}$ \quad Yunyang Ge$^{1}$\quad Mingyi Guo$^{1}$\quad \\
Li Yuan$^{1,2}$\textsuperscript{$\dag$} \quad Yonghong Tian$^{1,2}$\textsuperscript{$\dag$}\\ \\
$^{1}$Peking University \\
$^{2}$Pengcheng Laboratory \\
}

\begin{document}
\maketitle
\begin{strip}
	\centering
	\includegraphics[width=1.0\textwidth]{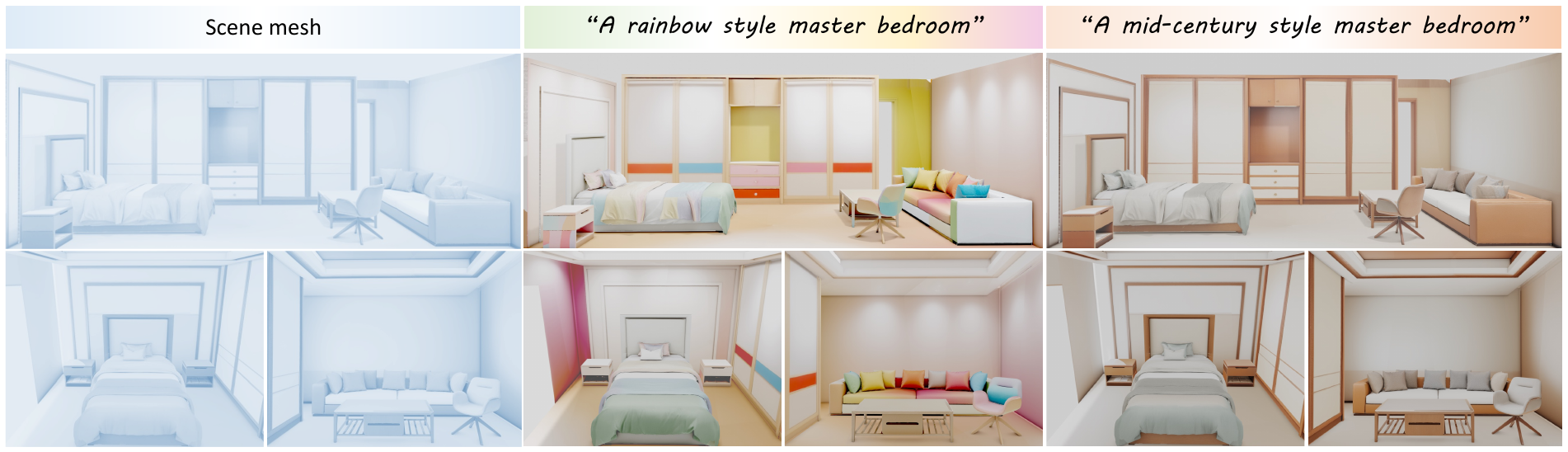}\\
	\captionsetup{type=figure,font=small}
	\caption{\textbf{RoomPainter} is capable of generating high-fidelity and consistent texture for a given room mesh.
	\label{fig:teaser}
    }
\end{strip}

\begin{abstract}
Indoor scene texture synthesis has garnered significant interest due to its important potential applications in virtual reality, digital media and creative arts. Existing diffusion-model-based researches either rely on per-view inpainting techniques, which are plagued by severe cross-view inconsistencies and conspicuous seams, or adopt optimization-based approaches that involve substantial computational overhead. In this work, we present \textbf{RoomPainter}, a framework that seamlessly integrates efficiency and consistency to achieve high-fidelity texturing of indoor scenes. The core of RoomPainter features a zero-shot technique that effectively adapts a 2D diffusion model for 3D-consistent texture synthesis, along with a two-stage generation strategy that ensures both global and local consistency. Specifically, we introduce Attention-Guided Multi-View Integrated Sampling \textbf{(MVIS)} combined with a neighbor-integrated attention mechanism for zero-shot texture map generation. Using the \textbf{MVIS}, we firstly generate texture map for the entire room to ensure global consistency, then adopt its variant, namely Attention-Guided Multi-View Integrated Repaint Sampling (\textbf{MVRS}) to repaint individual instances within the room, thereby further enhancing local consistency and addressing the occlusion problem. Experiments demonstrate that RoomPainter achieves superior performance for indoor scene texture synthesis in visual quality, global consistency and generation efficiency.
\end{abstract}    
\section{Introduction}
High-quality 3D content synthesis is in great demand, owing to its diverse applications across the entertainment industry, robotic simulation and mixed-reality environments. The design of indoor scene typically requires professionals to spend dozens of hours to complete. Generating texture for indoor scene through automated methods can significantly reduce labor costs and save time.
With the advancement of text-to-image diffusion models~\cite{Rombach_2022_sd, imagen, Ramesh2021ZeroShotTG, podell2023sdxl}, large-scale indoor scene texture synthesis has made significant progress.
Although existing indoor texture synthesis methods produce high-quality results, they face several significant challenges. Specifically, these methods can be categorized into two groups: inpainting-based methods~\cite{richardson2023texture, chen2023text2tex} and optimization-based methods~\cite{metzer2023latent-nerf, chen2024scenetex, youwang2024paint_it}.
Inpainting-based methods utilize depth-to-image diffusion models~\cite{Rombach_2022_sd, zhang2023controlnet} to progressively generate texture for neighboring views, with the guidance of reference geometry and previously generated views. However, due to their per-view inpainting strategy, these approaches often result in inconsistencies in texture style across instances within the room and fail to effectively address the loop-closure problem.
Optimization-based methods mitigate these inconsistencies by employing score distillation loss~\cite{poole2022dreamfusion} or its variations~\cite{wang2024prolificdreamer} in global texture maps. However, the optimization process incurs significant time costs for indoor scene texture synthesis, due to the introduction of a large number of learnable parameters and also prone to issue of training instability.
To address these challenges, we propose an efficient and high-fidelity indoor scene texturing framework, dubbed \textbf{RoomPainter}. 
The core of RoomPainter lies in a zero-shot technique that effectively adapts a 2D diffusion model for 3D-consistent texture synthesis, along with a two-stage generation strategy that ensures both global and local consistency. Specifically, we introduce Attention-Guided Multi-View Integrated Sampling \textbf{(MVIS)} combined with a related view-based attention mechanism for zero-shot texture map generation. Using the \textbf{MVIS}, we generate a holistic texture map for the entire room, achieving better global consistency than per-view inpainting-based methods. 
Subsequently, to paint the untextured areas raised by occlusions between objects while preserving the global consistency established in the first stage, we adopt a refine stage to repaint each instance in the room. In this stage, we apply an Attention-Guided Multi-View Integrated Repaint Sampling (\textbf{MVRS}) technique at the instance scale to perform texture inpainting and refinement for each individual instance.

The main contributions of our work are summarized below:
\begin{itemize}
    \item We propose \textbf{RoomPainter}, an effective framework for indoor scene texture synthesis that excels in producing consistent, high-quality texture with remarkable time efficiency.
   
    \item We introduce Multi-View Integrated Sampling (\textbf{MVIS}), a zero-shot technique that leverages view-weighted texture rectification and related view-based attention to generate global consistent texture map using 2D diffusion model.

    \item We further present Multi-View Integrated Repaint Sampling (\textbf{MVRS}), which simultaneously refines the textured areas and inpaints the occlusion areas based on their context, thereby facilitates fine-detailed texture synthesis with local consistency.

    \item We conduct extensive experiments and compare our method with strong baselines. The results demonstrate that our method not only achieves superior performance in visual quality for indoor scene texture synthesis but also excels in generation efficiency.
\end{itemize}
\begin{figure*}[!ht]
    \centering
    \includegraphics[width=\linewidth]{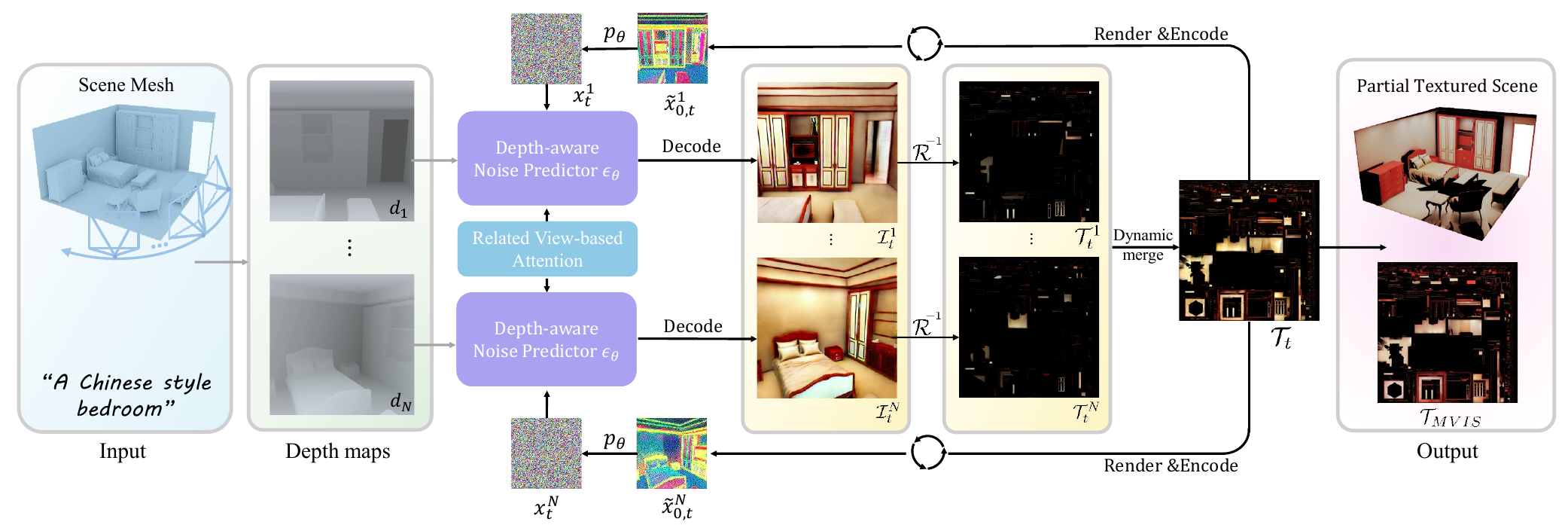}
    \caption{ 
    \textbf{Illustration of Multi-view Integrated Sampling}
        For $N$ surrounding viewpoints in the room, centered around the room's center point and guided by the corresponding depth maps, we use a diffusion model to generate the denoised observation $\mathcal{I}_t^n$ at timestep $t$. This observation is then projected into UV space to obtain texture maps that corresponding to respective viewpoints. The texture maps from multiple viewpoints are dynamically merged to produce the texture map for the current timestep, which subsequently guides the sampling process for the next timestep.
    }
    \label{fig:stage1_pipe}
\end{figure*}
\section{Related Works}
\subsection{3D Generation with Diffusion Model}
With the rapid advancements in image~\cite{Rombach_2022_sd,podell2023sdxl,zhang2023controlnet,mou2024t2i-adapter,imagen} and video generation models~\cite{svd,xing2025dynamicrafter,blattmann2023align,wu2023tune,yu2024viewcrafter,yu2025trajectorycrafter} in recent years, the field of 3D generation has experienced significant progress. Various methods now generate objects~\cite{poole2022dreamfusion,wang2024prolificdreamer,yu2024evagaussians,feng2025ae,yu2023hifi,yu2023nofa,wang2023score-3d} or scenes~\cite{hollein2023text2room,zhang2024text2nerf,ouyang2023text2immersion,zhou2024holodreamer,yu2024viewcrafter,cohen2023set_the_scene} based on textual descriptions via leveraging SDS Loss and its variants~\cite{poole2022dreamfusion,wang2024prolificdreamer} to optimize implicit representation such as NeRF~\cite{mildenhall2020nerf} and 3D Gaussian Splatting~\cite{kerbl20233d-gaussian}.
In the realm of scene generation, some approaches utilize text-guided inpainting models and depth prediction models, which incrementally construct scenes via a warping process~\cite{hollein2023text2room,fridman2024scenescape,zhang2024text2nerf,ouyang2023text2immersion,chung2023luciddreamer}. Other methods focus on fine-tuning image generation models using 3D data, allowing them to generate multi-view consistent images or panorama for 3D tasks with the guidance of semantic images or depth maps. Moreover, some methods predict depth from a given image to initialize 3D Gaussian Splatting, followed by further optimization of the 3D Gaussian Splatting using video generation models.

\subsection{3D Object Texturing}
The goal of 3D object texturing methods is to create high-quality texture for the given 3D object that align with user-provided textual descriptions. Early methods, such as~\cite{richardson2023texture,chen2023text2tex,tang2024intex,wang2024roomtex}, utilized depth-aware Diffusion Models~\cite{Rombach_2022_sd,zhang2023controlnet,mou2024t2i-adapter} to iteratively apply inpainting for each viewpoint. At each step, the texture is generated based on the texture from the previous viewpoint, gradually rotating the object to produce a complete texture. While these methods can generate texture in a relatively short time, they fail to address issues such as loop-closure and visible seams.
Other works, such as~\cite{chen2023fantasia3d,metzer2023latent-nerf,youwang2024paint_it}, leverage differentiable rendering combined with SDS~\cite{poole2022dreamfusion} to texture the given object through optimization techniques.
Additionally, approaches like~\cite{cao2023texfusion,liu2023syncmvd,gao2024genesistex,huo2024texgen} use UV maps as intermediaries and apply different fusion methods to integrate multiple of view independent UV maps into a consistent UV map. During the Diffusion model sampling process, this consistent UV map is used to adjust the Diffusion model's input, ensuring that the resulting texture is more consistent, with fewer visible seams and improved multi-view consistency.

\subsection{3D Indoor Scene Texturing}
In the task of 3D indoor scene texture generation, the objective is to create texture for a given indoor scene based on a user-provided textual description, which presents challenges such as object occlusion and multi-view consistency. SceneTex~\cite{chen2024scenetex} defines an implicit texture field and optimizes it using VSD~\cite{wang2024prolificdreamer}. By employing an optimization-based approach to refine the texture field, SceneTex naturally resolves the issue of multi-view consistency in 3D scene texture. Furthermore, SceneTex incorporates a cross-attention mechanism to transfer texture from visible regions to occluded areas. However, this process can result in blurriness in the occluded regions, and the optimization procedure itself is computationally intensive.
When methods like Text2Tex and TEXTure~\cite{chen2023text2tex,richardson2023texture}, typically used for object texture generation, are applied to scene texture generation, they can only enforce style consistency across multiple objects based on textual control. However, these methods do not ensure global style consistency for the entire scene.
Alternatively, if the entire room is treated as a single object and methods such as TEXTure or Text2Tex are employed with the camera positioned at the center of the room, generating the scene texture by rotating the camera, global style consistency can be achieved. However, this approach still faces challenges such as occlusion and visible seams between regions.

Other approaches, such as~\cite{yang2024dreamspace}, use panoramic images to generate a rough texture for the entire room. The camera position is then changed to generate texture for occluded regions. Subsequently, an MLP is employed to predict the texture of unseen areas based on the color and texture coordinates from the UV map. However, this method requires the scene to have an initial texture and relies on texture style transfer, meaning it cannot generate texture from scratch for a completely untextured scene.

\section{Methods}
Given a 3D indoor scene mesh $\mathcal{M}$, our objective is to efficiently generate texture that not only align with the provided textual descriptions but also maintain global consistency.
Since our framework employs pretrained text-to-image diffusion model to generate texture for the given indoor scene mesh, we will start with an overview of the text-to-image Diffusion model and mesh rendering in Sec.~\ref{subsec:preliminaties}. 
Following this, we introduce Multi-View Integrated Sampling (abbreviated as \textbf{MVIS}) in Sec.~\ref{subsec:MVIS}, which addresses the challenge of maintaining multi-view consistency while preserving high-frequency details in the diffusion model sampling process. 
Subsequently, In Sec.~\ref{subsec:MVRS}, we introduce a variant of \textbf{MVIS}, namely Multi-View Integrated Repaint Sampling (\textbf{MVRS}), to inpaint the occlusion region and simultaneously refine existing texture for each instance within the indoor scene.
Finally, in Sec.~\ref{subsec:RV-attn}, we introduce a Related View-based Attention mechanism used in the \textbf{MVIS} and \textbf{MVRS} processes, which enhances multi-view consistency during texture synthesis.

\subsection{Preliminaries}
\label{subsec:preliminaties}

\noindent\textbf{Diffusion Model Sampling.} Diffusion models~\cite{ho2020denoising,sohl2015deep} are latent variable models that comprises a forward process $q(x_t|x_0)$ and a learned sampling process $p_\theta(x_{t-1}|x_t)$. Given an initial data point $x_0$, its noisy counterpart at time t is denoted as $x_t$, and the forward process $q$ is expressed as:
\begin{equation}
\label{diffsuion_q}
x_t = \sqrt{\bar\alpha_t}x_0 + \sqrt{(1-\bar\alpha_t)}\epsilon, ~\epsilon~\sim N(0, I),
\end{equation}
where $t \in [0, \infty)$ and the values of $\alpha_t$ are defined by a scheduler to controls the amount of noise added to $x_0$ at time step $t$. Additionally, it has $\beta_t=1-\alpha_t$ and $\bar\alpha_t=\prod_{s=1}^{t} \alpha_s$. 
In the sampling process $p_\theta$, with the DDPM sampler~\cite{ho2020denoising}, the diffusion model progressively denoises an initial noise image $x_T\sim \mathcal{N}(0,I)$ to clean image. Specifically, at any intermediate time step $t$, the sampling process $p_\theta$ can be formed as: 
\begin{equation}
\label{eq:diffusion_prev}
x_{t-1}=\mu_{t-1} + \sigma_t \epsilon, ~~\epsilon\sim N(0,I),
\end{equation}
where the value of $\sigma_t$ and $\mu_{t-1}$ is given by:
\begin{equation}
\label{eq:diffusion_mu}
\quad \mu_{t-1} \ {=} \ \frac{\sqrt{\bar\alpha_{t-1}}\beta_t}{1-\bar\alpha_t} x_{0, t} + \frac{\sqrt{\alpha_t}(1- \bar\alpha_{t-1})}{1-\bar\alpha_t}x_t,
\end{equation}
\begin{equation}
\label{eq:diffusion_sigma_t}
\sigma_t = \frac{1-\bar\alpha_{t-1}}{1-\bar\alpha_t}\beta_t,
\end{equation}
and the estimation of the original image $x_{0,t}$ at time step $t$ is calculated as: 
\begin{equation}
\label{eq:diffusion_pred_x0}
x_{0, t} = \dfrac{x_t - \sqrt{1 - \bar\alpha_t}\epsilon_\theta(x_t,t)}{\sqrt{\alpha_t}},
\end{equation}
where $\epsilon_\theta(x_t,t)$ is the noise that estimated by the diffusion model.
As $t$ decreases from $T$ to $0$, the sampling process $p_\theta$ ultimately samples a clean data point $x_0$.
Moreover, this process can be generalized for learning a marginal distribution using an additional input condition. That leads depth-aware text-to-image diffusion models~\cite{controlnet-depth, zhang2023controlnet}, where the output of the model $\epsilon_\theta(x_t,y,d,t)$ is conditioned on a text prompt $y$ and a depth map $d$.

In this paper, we utilize Stable Diffusion~\cite{podell2023sdxl} as our base model. It performs denoising in the latent space and employs an autoencoder $\mathcal{D(\mathcal{E(\cdot)})}$ for the conversion between image and latent representations. Therefore, latent $x_0$ generated by its diffusion process will be decoded to the image $\mathcal{D}(x_0)$ through its decoder.

\noindent\textbf{Mesh Rendering.} Given a mesh $\mathcal{M}$, a texture map $\mathcal{T}$ and a viewpoint $C$, the rendering function $\mathcal{R}$ can be employed to produce the rendered image as $I = \mathcal{R}(\mathcal{T}, \mathcal{M}, C)$. Conversely, the inverse rendering function: $\mathcal{R}^{-1}$ can reconstruct the texture map from the image, yielding $\mathcal{T}^{\prime} = \mathcal{R}^{-1}(I, \mathcal{M}, C)$.

\subsection{Mult-View Integrated Sampling}
\label{subsec:MVIS}

Different from object-level multi-view diffusion models~\cite{shi2023mvdream,liu2023zero123,shi2023zero123++}, diffusion models that capable of ensuring multi-view consistency in scene level image generation~\cite{Tang2023mvdiffusion} are extremely limited. 

To address this challenge, we propose Multi-View Integrated Sampling (\textbf{MVIS}), grounded with a Related View-based Attention technique, to adapt a standard text-to-image model to generate view-consistent images in a training-free manner.
Based on \textbf{MVIS}, we first implement a global texturing stage to generate consistent texture for the scene mesh. 

As shown in Fig.~\ref{fig:stage1_pipe}, given a set of predefined cameras $\{C\} = C^n, n=1,\dots,N$, we first render their corresponding depth maps $d_n$ and similarity masks $S_n$, where each pixel in the similarity mask represents the inverted normalized value of the cosine similarity between the normal vectors of the visible faces and the viewing direction, with values ranging from $[0,1]$~\cite{richardson2023texture,chen2023text2tex,gao2024genesistex}. 
Based on the similarity masks, we construct the texture weighting maps $W_n$ for each camera, defined as $W_n = \mathcal{R}^{-1}(S_n, \mathcal{M}, C^n)$, which are then used to dynamically merge per-view texture maps.

As shown in Algorithm.~\ref{alg:MVIS}, in the sampling process, we firstly sample $\{x_T^{n}\}_{n=1}^N\sim \{\mathcal{N}({0}, {I})\}$ for $N$ camera views. 
Subsequently, at each denoising step $t$, we decode all $N$ estimated $x_{0, t}^n$ into $\mathcal{I}_t^n$ and back project them into independent per-view texture map $\mathcal{T}_t^n$. We then merge these per-view texture maps to form a global texture map $\mathcal{T}_t$ using dynamic merge~\cite{liu2023syncmvd,bensadoun2024meta}, which can be formulated as:
\begin{equation}
\label{eq:dynamical_merge}
\mathcal{T}_t=\frac{\sum_{n=1}^N (W_n^{exp(t)} \cdot \mathcal{T}_t^n)}{\sum_{n=1}^N  W_n + \gamma}.
\end{equation}
Here, $exp(t)$ is used to dynamically regulate variations in consistency during the sampling process at timestep $t$. As $t$ decreases, $exp(t)$ increases linearly, leading to a sharper merged texture.
With the merged texture map, we again render per view images from $\mathcal{T}_t$, using foreground mask $M^n$ for viewpoint $C^n$, and encode them into latent space to obatain $\tilde x_{0, t}$, which is then used to replace the previous obtained less $x_{0, t}^n$ in Eq.~\ref{eq:diffusion_mu} for consistent view generation. After repeating these denoising steps, we are able to generate a consistent texture map, denote as $\mathcal{T}_{MVIS}$.

\begin{algorithm}[t]
\caption{Multi-View Integrated Sampling}\label{alg:MVIS}
\begin{algorithmic}
\STATE Input: mesh $\mathcal{M}$, text $y$, cameras $\{C^1, \dots, C^N \}$
\STATE Parameters: DDPM noise schedule $\{\sigma_t\}_{t=T}^0$
\STATE Initialization: $\{x_T^{n}\}_{n=1}^N\sim \{\mathcal{N}({0}, {I})\}$ 
\FOR{$t \in \{T \dots 0 \}$} 
    \FOR{$n \in \{1 \dots N\}$}
        \STATE ${\epsilon_t^n} \leftarrow \epsilon_\theta(x_t^n, y, d_n, t)$ 
        \STATE $ x_{0, t}^n \leftarrow \dfrac{x_t^n - \sqrt{1 - \bar\alpha_t} {\epsilon_t^n}} {\sqrt{\alpha_t}}$ 
        \STATE $\mathcal{{I}}_t^n \leftarrow \mathcal{D}( x_{0, t}^n )$ 
        \STATE $\mathcal{T}_t^n \leftarrow \mathcal{R}^{-1}( \mathcal{I}_t^n, \mathcal{M}, C^n )$ 
    \ENDFOR \vspace{2pt}

    \STATE $\mathcal{T}_t = dynamic\_merge(\{\mathcal{T}_t^n\}_{n=1}^N)$ \vspace{2pt}
    
    \IF{ $t > 0$ }
        \FOR{$n \in \{1 \dots N\}$} 
            \STATE $\epsilon^n \sim \mathcal{N}(\mathrm{0}, \boldsymbol{I})$ 
            \STATE ${\tilde x_{0, t}^n} \leftarrow \mathcal{E}( {M^n} \odot {\mathcal{R}(\mathcal{T}_t, \mathcal{M}, C_n)} + {(1-M^n)} \odot {\mathcal{I}_t^n} )$ 
            \STATE $ {\mu_{t-1}^n} \leftarrow \frac{ {\sqrt{ \bar\alpha_{t-1} } } {\beta_t} } {1-\bar\alpha_t} {\tilde x_{0, t}^n} + \frac{\sqrt{\alpha_t}(1- \bar\alpha_{t-1})}{1-\bar\alpha_t}x_t^n$ 
            \STATE $x_{t-1}^n \leftarrow \mu_{t-1}^n~+~\sigma_t\epsilon^n$ 
        \ENDFOR
    \ENDIF
\ENDFOR

$\mathcal{T}_{MVIS} = \mathcal{T}_0$
\RETURN Texture map $\mathcal{T}_{MVIS}$
\end{algorithmic}
\end{algorithm}


\subsection{Multi-View Integrated Repaint Sampling}
\label{subsec:MVRS}
\begin{figure*}[!ht]
    \centering
    \includegraphics[width=\linewidth]{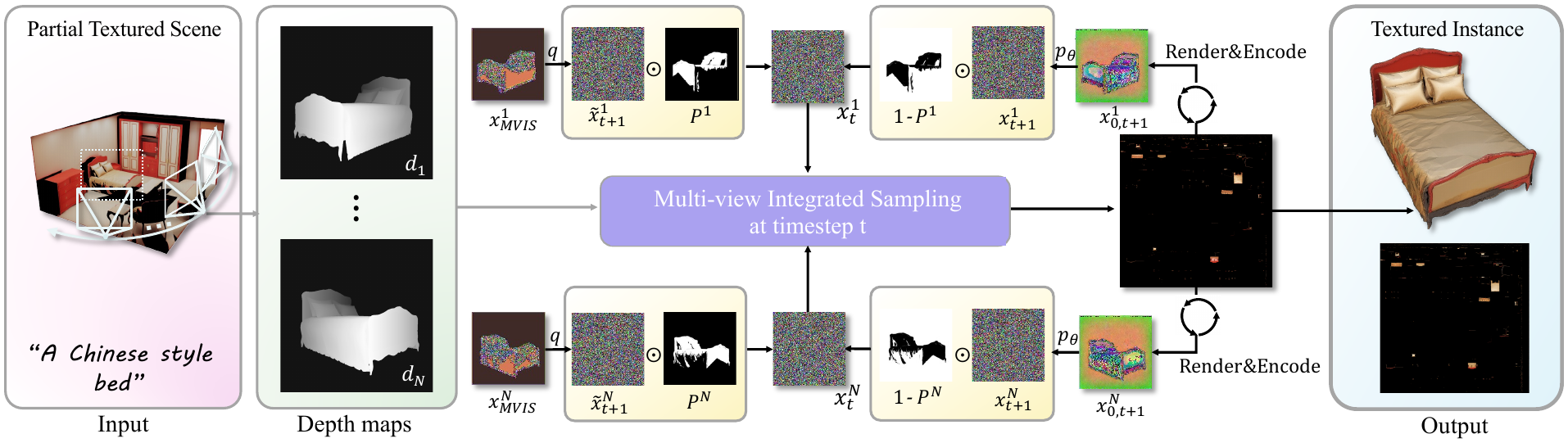}
    \caption{ 
    \textbf{Illustration of Multi-View Integrated Repaint Sampling(MVRS)}
        Due to occlusion between instances, certain areas remain untextured after the first stage of texture generation. To address this issue, we perform texture generation for each instance within the room based on the painted areas. For $N$ different viewpoints of each instance, guided by the corresponding depth maps, at sampling step $t$ of MVRS, we combine the painted areas($x_{MVIS}$) with the sampling results of MVIS at step $t+1$ ($x_{0,t+1}$) using a mask $P$ to form $x_t$, which serves as the input for the sampling process of timestep t. Noise corresponding to the current timestep was added before mask combine. Upon completing the MVRS process, the texture map for a specific instance in the room is fully generated. 
        }
    \label{fig:stage2_pipe}
\end{figure*}

Although the texture map $\mathcal{T}_{MVIS}$ maintains stylistic and consistent throughout the room, there are still areas that remain untextured due to occlusions. 
To address this problem, we adopt a refine stage to inpaint the occlusion areas and further enhance the texture fidelity, while preserving the global consistency established in the first stage.

Specifically, we decouple each instance in the room, resulting a total of $K$ instances $\mathcal{M}_1, \dots, \mathcal{M}_K$. We then design a variant of \textbf{MVIS}, namely multi-view integrated repaint sampling (\textbf{MVRS}), to perform texture inpainting and refinement for each individual instance. 
As shown in Fig.~\ref{fig:stage2_pipe}, the core of \textbf{MVRS} lies in incorporating diffusion-based inpainting~\cite{lugmayr2022repaint} into the \textbf{MVIS} process. For each individual instance, given $N$ predefined camera views, we firstly render $N$ images $\mathcal{I}^{n}$ along with their inpainting masks $P^n$ that indicates untextured region of them. Subsequently, for any time step $t$ in the \textbf{MVIS} process, we encode $\mathcal{I}^{n}$ into the latent space and add noise on it to produce a noisy latent $\tilde{x}_{t}^n$, followed by sampling ${x}_{t}^n$ from \textbf{MVIS}, then combine them into a new latent using the inpainting mask $P^n$, where the painted area keep unchanged, finally passing it into the U-Net for the next denoising step. 
This process introduces information from the painted area into the \textbf{MVIS} process, effectively transferring the texture of the painted area to the regions with holes while ensuring consistency across multiple viewpoints.
After iterating through all instances within the room, we obtain the final output scene texture $\mathcal{T}_{MVRS}$.

\subsection{Related View-based Attention}
\label{subsec:RV-attn}
In the \textbf{MVIS} and \textbf{MVRS} process, we employ a 2D Diffusion Model for zero-shot multi-view sampling. A key challenge in maintaining view consistency is sharing information across multiple images from different viewpoints. Inspired by~\cite{hertz2024style-aligned}, we modify the self-attention mechanism in the diffusion model to ensure that each view incorporates sampling information from related views during the sampling process. 
During the sampling process, denote the queries, keys and values derived from the deep feature of the denoising U-Net for viewpoint $n$ as $Q_n$, $K_n$, and $V_n$, respectively. For each viewpoint, assume that it has a total of $R$ associated viewpoints (view that has overlap). Subsequently, the Related View-based Attention for viewpoint $n$ is computed as:
\begin{equation}
    \label{eq:Attention}
    softmax\left( \dfrac{Q_n \tilde K_n^T}{\sqrt{d}} \tilde V_n \right),
\end{equation}
where $\tilde K_n=\begin{bmatrix} K_1  \ K_{2}  \ \hdots \ K_{R} \end{bmatrix}^T$ 
and $\tilde V_n=\begin{bmatrix} V_1  \ V_2  \ \hdots \ V_{R}  \end{bmatrix}^T$. 
In the first room-scale texturing stage, we consider the $R$ related viewpoints for view $n$ as its adjacent views, \textit{i.e.}, its left and right view. When sampling at instance scale, we define the related views as all the $N$ predefined camera view. 
\section{Experiments}
In this section, we first describe the general implementation of the baseline methods used for comparison. Next, we provide an overview of the implementation details of our proposed method. Finally, we present our quantitative experiments, qualitative results, and ablation studies.
\begin{table}[!t]
    \centering
    \resizebox{\linewidth}{!}{
        \begin{tabular}{l c cc }
            \toprule
            \multirow{1}{*}{Method} & Generation Time(mins)~$\downarrow$ & CLIP~$\uparrow$ & AS~$\uparrow$ \\
            \midrule
            Text2Tex-H~\cite{chen2023text2tex} & \textbf{8.50} & 21.58 & 4.34  \\
            Text2Tex-C~\cite{chen2023text2tex} & 70.75 & 21.93 & 4.85  \\
            SceneTex~\cite{chen2024scenetex} & 2614.50 & 21.87 & 4.75  \\
            (Ours) Full & 46.00 & \textbf{23.47} & \textbf{5.03}  \\
            \midrule
             (Ours) w/o Attn & - & 23.32 & 5.01  \\
             (Ours) w/o MVIS & - & 23.27 & 5.01  \\
             (Ours) w/o MVRS & - & 22.39 & 4.40  \\
             
            \bottomrule
        \end{tabular}
    }
    \caption{
        \textbf{Quantitative comparisons.}
        We report the Generation time and 2D metrics result for quantitative comparisons. The generation time is shown in the format minutes. The 2D metrics includes: CLIP Score (CLIP)~\cite{hessel2021clipscore}, Aesthetic Score(AS)~\cite{schuhmann2021laion}. We show that our method produces high quality textures efficiently.
    }
    \label{tab:quantitatives}
\end{table}
\begin{figure*}[ht]
    \centering
    \includegraphics[width=\linewidth]{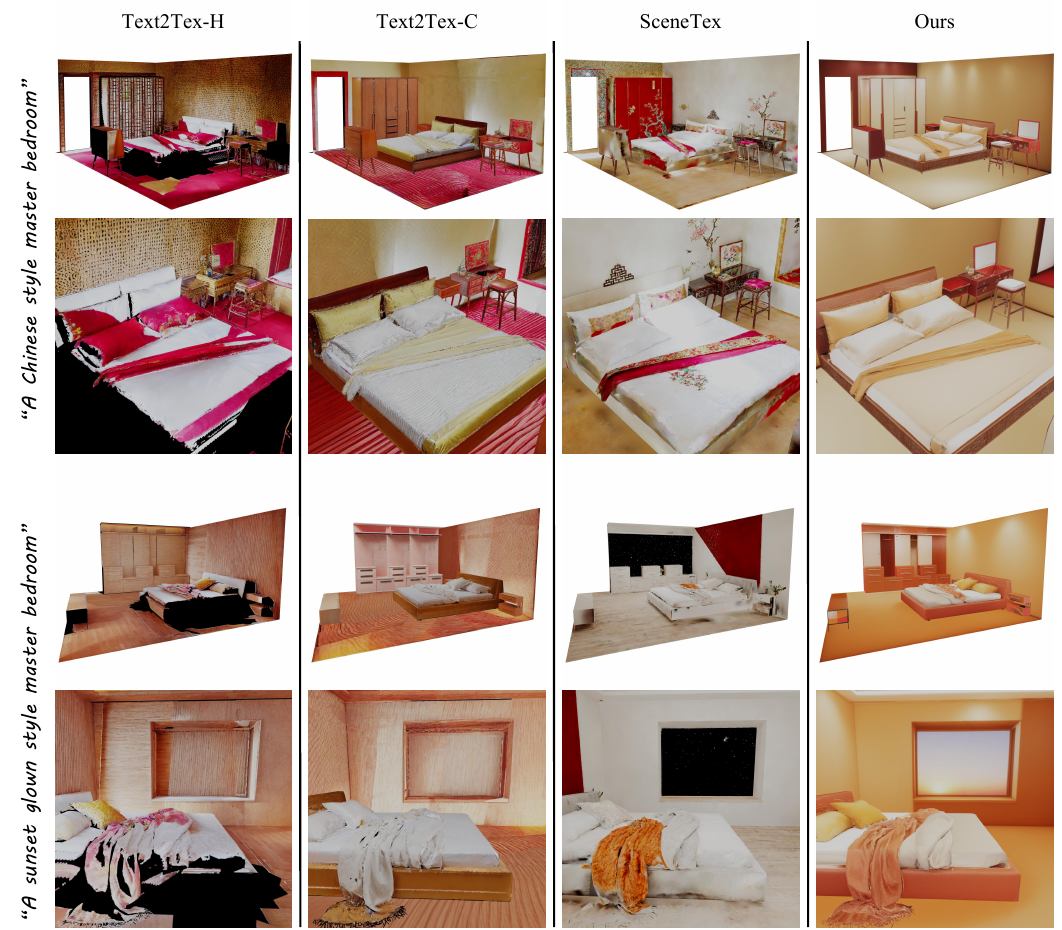}
    \caption{ 
    \textbf{Qualitative comparisons.}
        Text2Tex-H~\cite{chen2023text2tex} suffers from occlusion and visible seams. Text2Tex-C~\cite{chen2023text2tex} struggles to maintain style consistency across all instances. SceneTex~\cite{chen2024scenetex} produces unrealistic texture and results in blurry regions. In contrast, our method generates high-quality texture while preserving overall style consistency across instances in the scene. Ceilings and back-facing walls are excluded for improved visualizations.
    }
    \label{fig:comp_baselines}
\end{figure*}
\begin{figure*}[t]
    \centering
    \includegraphics[width=\linewidth]{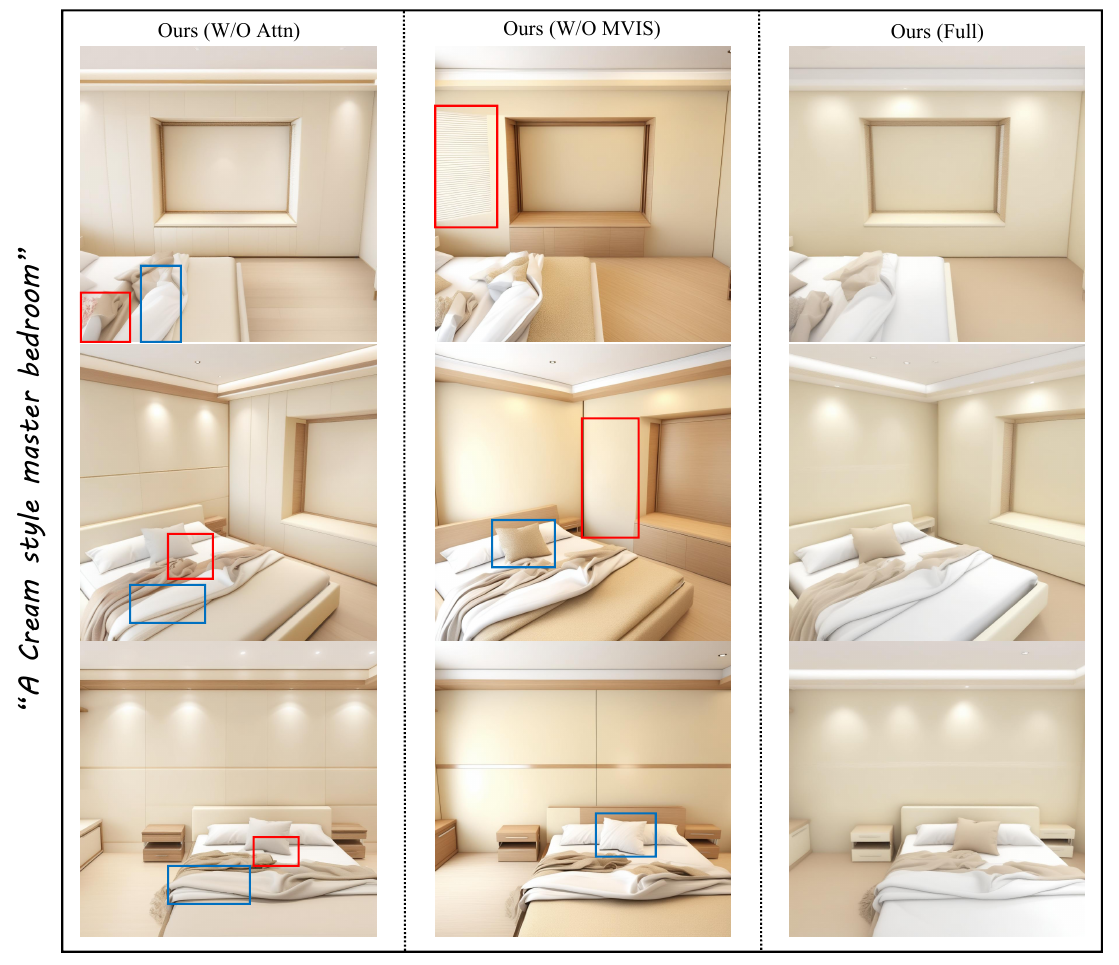}
    \caption{ 
    \textbf{Ablation studies on the multiview-consistency module.}
        Synthesizing texture without the Related View-based attention leads to inconsistencies across views, as shown in the leftmost column. When synthesizing texture without MVIS, noticeable inconsistencies appear across different viewpoints (middle column). In contrast, our full method samples multi-view images with significantly stronger consistency. The areas of major inconsistency in the images are highlighted with red and blue boxes. In the same column, boxes of the same color indicate that they are close to each other in 3D space. Zoom in for the best view. 
    }
    \label{fig:ablation_module_fig}
\end{figure*}

\subsection{Baseline Implementation Details}
\begin{itemize}
    \item{Text2Tex-H}~\cite{chen2023text2tex}: This variant of Text2Tex is implemented using a holistic text prompt provided by user and treats the entire room as a cohesive entity. The camera is positioned at the center of the room, with a field of view of $60^\circ$, and the elevation angle is set to $0^\circ$, and azimuth angles uniformly sampled from the range $[0^\circ, 360^\circ]$, which results in a total of twelve perspectives.
    \item{Text2Tex-C}~\cite{chen2023text2tex}: For comparative purposes, we employ the original configuration of Text2Tex, wherein each room instance, including the interior walls and furniture, is textured using an instance-specific text prompt. And for interior walls texturing, we utilize the text prompt such as: "A Chinese style living room, no furniture" and sampled 12 perspectives same as Text2Tex-H. All the textured instances are subsequently combined, forming the output textured room.
    \item{SceneTex}~\cite{chen2024scenetex}: We compare against SceneTex using its original setup, while employing the same holistic prompt as utilized in our approach.
\end{itemize}

\subsection{Implementation Details}
\label{subsec:implent_detail_ours}
In this experiment, we used the SDXL model~\cite{podell2023sdxl} as the text-to-image generator and incorporated depth information through ControlNet~\cite{zhang2023controlnet,controlnet-depth}. For the text prompt, we implemented view-specific prompt combining the holistic prompt with the names of the types of objects present within the field of view. For sampling, we applied the DDPM~\cite{ho2020denoising} as the sampler, setting the sampling steps to 50. We used PyTorch3D~\cite{ravi2020pytorch3d} for rendering and texture projection. Additionally, UV atlas generation for the mesh was handled with Xatlas~\cite{xatlas2016}. All experiments were performed on an NVIDIA RTX A6000 GPU. Further implementation details are available in the supplementary materials.

\subsection{Quantitative Analysis}
Our quantitative experiment were conducted across 10 scenes including 5 bedroom and 5 living room from the 3D-FRONT dataset~\cite{fu20213d-front}, using five different text styles totally.
To evaluate the generated texture, we calculated both the CLIP Score(CLIP)~\cite{hessel2021clipscore}, which assesses the alignment between generated texture and the textual description, and the CLIP Aesthetic Score(AS)~\cite{schuhmann2021laion}, which measures texture quality. We also compared the average processing time of each texture generation method in minutes. The results are summarized in Table.\ref{tab:quantitatives}. We render 10 images from novel views for each room to calculate the average of CLIP and AS. 
Due to the presence of numerous distinct objects within the room, the methods in baselines do not adjust the text prompts for texture generation based on the variation of objects from different viewpoints. Therefore, we employ a holistic prompt to compute the CLIP score. 
Since our evaluation metric involves image sampling from new viewpoints, Text2Tex-H, which fails to address the occlusion problem, generates texture containing large occluded regions, resulting in the lowest performance in terms of the metric. However, due to the limited number of sampled viewpoints, it exhibits the fastest generation speed.
We configure SceneTex according to its original setup, sampling from a sphere centered at the room’s center. As a result, regions not observed by SceneTex often exhibit blurriness. 
Meanwhile, in addition to optimizing the texture field using VSD, SceneTex also requires training LoRA, which results in the longest processing time. 
Text2Tex-C employs an instance-specific approach for texture generation, utilizing different prompts for each object, which results in higher performance metrics. Additionally, it incorporates viewpoint selection and texture optimization processes, leading to a longer computation time.

\subsection{Qualitative Analysis}
We provide a visual comparison in fig.\ref{fig:comp_baselines}. Specifically, Text2Tex-H~\cite{chen2023text2tex} shows that simply applying object texture generation method to indoor scene texturing task cannot resolve the occlusion problem (see the large black areas in the first column of fig.\ref{fig:comp_baselines}) and contain obvious seams. Text2Tex-C~\cite{chen2023text2tex} texturing the room completely but suffering from obvious seams and global style consistency across objects. Although SceneTex introduces the use of cross-attention to generate texture for occluded regions, noticeable blurriness still persists in these areas. Furthermore, due to its optimization-based approach, SceneTex often produces unrealistic texture. In contrast, our method generates globally consistent texture with a unified style in the first stage. In the second stage, through per-instance inpainting and refinement, we successfully maintain global style consistency while addressing issues of occlusion and unrealistic texture.

\subsection{Ablation Studies}
As shown in Fig.\ref{fig:ablation_module_fig}, all images are raw samples from the final step, thus revealing inherent multi-view inconsistencies. 

\noindent\textbf{Effectiveness of the MVIS.} 
In the first stage of texturing, we introduce MVIS to enhance the global consistency of the texture. Qualitative results from this stage are shown in the second column of Fig.\ref{fig:ablation_module_fig}. The figure demonstrates that without MVIS, texture consistency across multiple viewpoints can be significantly impact. We further conduct quantitative ablation on dataset adopted in previous experiments. Results are shown in the seventh row of Tab.\ref{tab:quantitatives}. The quantitative experimental results clearly demonstrate that MVIS improves the quality of the generated texture.  

\noindent\textbf{Effectiveness of the Related View-based Attention.} 
We incorporate the Related View-based Attention module in both stages of texture generation to enhance consistency. Both qualitative(the first column of Fig.\ref{fig:ablation_module_fig}) and quantitative experiments(the fifth row of the Tab.\ref{tab:quantitatives}) demonstrate that using only MVIS or MVRS is insufficient to generate consistent and high-quality texture. The Related View-based Attention module effectively enhances consistency while ensuring the quality of the generated texture.

\noindent\textbf{Effectiveness of the MVRS.} 
In the second stage of texturing, we use MVRS to generate texture for the occluded regions of each instance and refine the texture generated in the first stage. Quantitative experimental results (seventh row in Tab.\ref{tab:quantitatives}) show that without MVRS, the quality of the texture significantly deteriorates. This is because, without MVRS, the texture generated in the first stage suffer from low-quality results due to occlusion. We will provide additional qualitative materials demonstrating the effectiveness of MVRS in the supplementary materials.
\section{Conclusion}
We propose an indoor scene texturing framework named RoomPainter that generates high-quality texture with multi-view consistency and time efficiency.
By leveraging the carefully designed modules including view-weighted texture rectification and related view-based attention, proposed view-integrated diffusion preserves the cross-view consistency during the diffusion process.
Qualitative and quantitative results demonstrate the effectiveness of RoomPainter in indoor scene texturing, especially in global consistency and generation efficiency.
{
    \small
    \bibliographystyle{ieeenat_fullname}
    \bibliography{main}
}
\clearpage
\appendix
\section*{Supplementary Material}
\section{Implementation Details}
\subsection{Text Prompt of Stage1}
In the first stage of texture generation, we use different text prompts when sampling the image corresponding to each view. To determine the text prompt for a specific view, we calculate the proportion of pixels rendered for each instance in the view relative to the total number of pixels. If this proportion exceeds a set threshold $\sigma$, we include the text corresponding to that instance in the global prompt. For example, taking \textbf{``A Chinese style bedroom''} as the holistic prompt, if the proportions of instances: \textbf{``single bed'', ``wardrobe'' and ``chair''} exceed $\sigma$, the text prompt used for sampling this view becomes \textbf{``A Chinese style bedroom with single bed, wardrobe and chair.''} In our implementation, the threshold $\sigma$ is set to $0.01$.

\subsection{Text Prompt of Stage2}
During the texture generation phase for each instance, we derive the style prompt from the holistic prompt. For example, in \textbf{``A Chinese style bedroom''}, the style prompt is \textbf{``Chinese''}. When applying MVRS to a room without furniture, we use the same prompt for all viewpoints, such as \textbf{``A Chinese style bedroom, without furniture.''} For individual furniture instances, we use view-specific prompts. For example, for the instance \textbf{``single bed''}, the prompt would be \textbf{``A Chinese-style single bed, [DIR] view''} where [DIR] represents the relative position of the viewpoint. The [DIR] value changes based on the azimuth angle of the viewpoint and may include directions such as `front', `front side', `rear', `side', or `top-down'.

\subsection{Time steps for MVIS}
During the MVIS sampling process, we adopt different strategies based on the time step $t$. When $t \in \{T \dots 0.9T \}$, we use the standard diffusion sampling method. For $t \in \{0.9T \dots 0.5T \}$, the MVIS is applied. During $t \in \{0.5T \dots 0.3T \}$, we alternate between diffusion sampling and the MVIS. Finally, for $t \in \{0.3T \dots 0\}$, we revert to the standard diffusion sampling method. This multi-stage sampling strategy accelerates the overall sampling process while maintaining high texture quality.
\begin{table}[!t]
    \centering
    \setlength{\tabcolsep}{10pt}
    \resizebox{1.0\linewidth}{!}{
        \begin{tabular}{l c c c c}
            \toprule
            \multirow{1}{*}{Method} & PF~$\uparrow$ & VQ~$\uparrow$ & TG~$\uparrow$ \\
            \midrule
            Text2Tex-H~\cite{chen2023text2tex} & 3.06 & 2.72 & 2.43  \\
            Text2Tex-C~\cite{chen2023text2tex} & 3.22 & 3.10 & 3.15  \\
            SceneTex~\cite{chen2024scenetex} & 3.47 & 3.28 & 3.42  \\
            Ours & \textbf{4.00} & \textbf{4.13} & \textbf{4.31}  \\
            \bottomrule
        \end{tabular}
    }
    \caption{
        \textbf{User study result.}
        We report the prompt fidelity (PF), visual quality (VQ) and texture-geometry alignment (TG) results for user study. We show that our method produces high quality textures that coherent with input text prompt.
    }
    \label{tab:user_study}
\end{table}
\subsection{Time steps for MVRS}
Similar to the multi-stage sampling strategy in MVIS, we also adopt a multi-stage approach in MVRS to accelerate the process and improve texture quality. When $t \in \{T \dots 0.8T \}$, we use the standard RePaint sampling method without projecting images to texture space. For $t \in \{0.8T \dots 0.5T \}$, we apply the standard MVRS method. During $t \in \{0.5T \dots 0.3T \}$, we alternate between diffusion sampling and the MVIS. Finally, for $t \in \{0.3T \dots 0\}$, we switch back to the standard diffusion sampling method.

\subsection{Camera selection of both stage}
Our two-phase camera configuration:

\noindent{Global Phase}: N cameras (N=6) are positioned to look at the center of the room, with a radius equal to half the shortest axis of the room.

\noindent{Instance Phase}: N cameras (N=12 for room-frame repainting, N=9 for furniture repainting) are positioned around each instance, looking at its center. The camera distance is set to 0.95 times of the diagonal length of the furniture's bounding box when repainting furniture. For room-frame repainting, the same camera parameters as global phase are used, but with a field of view (FOV) of 80 degrees.

\subsection{Parameters}
Throughout all sampling processes, we set the inference steps to $50$. The classifier-free guidance scale is linearly reduced from $10.0$ to $7.0$ over the sampling time steps. The ControlNet conditioning scale is fixed at 1. Additionally, the exp parameter in $dynamic\_merge$ is linearly increased from $1.0$ to $6.0$ over the sampling time steps. The $T$ value is consistent with the default setting in Stable Diffusion, which is $1000$.

\section{A more intuitive explanation of the second stage.}
The detailed implementation of MVRS is presented in Alg.\ref{alg:MVRS}. After completing the first stage of texture generation for the entire room, an initial texture $\mathcal{T}_{MVIS}$ is obtained. However, due to occlusion issues, the initial texture contains numerous untextured black regions. In MVRS, for the n-th viewpoint, the regions with an initial texture are denoted as $P^n$, and the corresponding rendered image is represented as $\mathcal{I}_{MVIS}^n$.
To address the occlusion problem, we perform separate MVRS operations for each instance in the room. After completing the MVRS operations for all instances, we take the union of the texture maps corresponding to each instance to produce the final texture map for the entire scene. This final texture represents the output generated by our proposed framework.

\section{Ablation study on the scenond stage}
\begin{figure*}[!b]
    \centering
    \includegraphics[width=\linewidth]{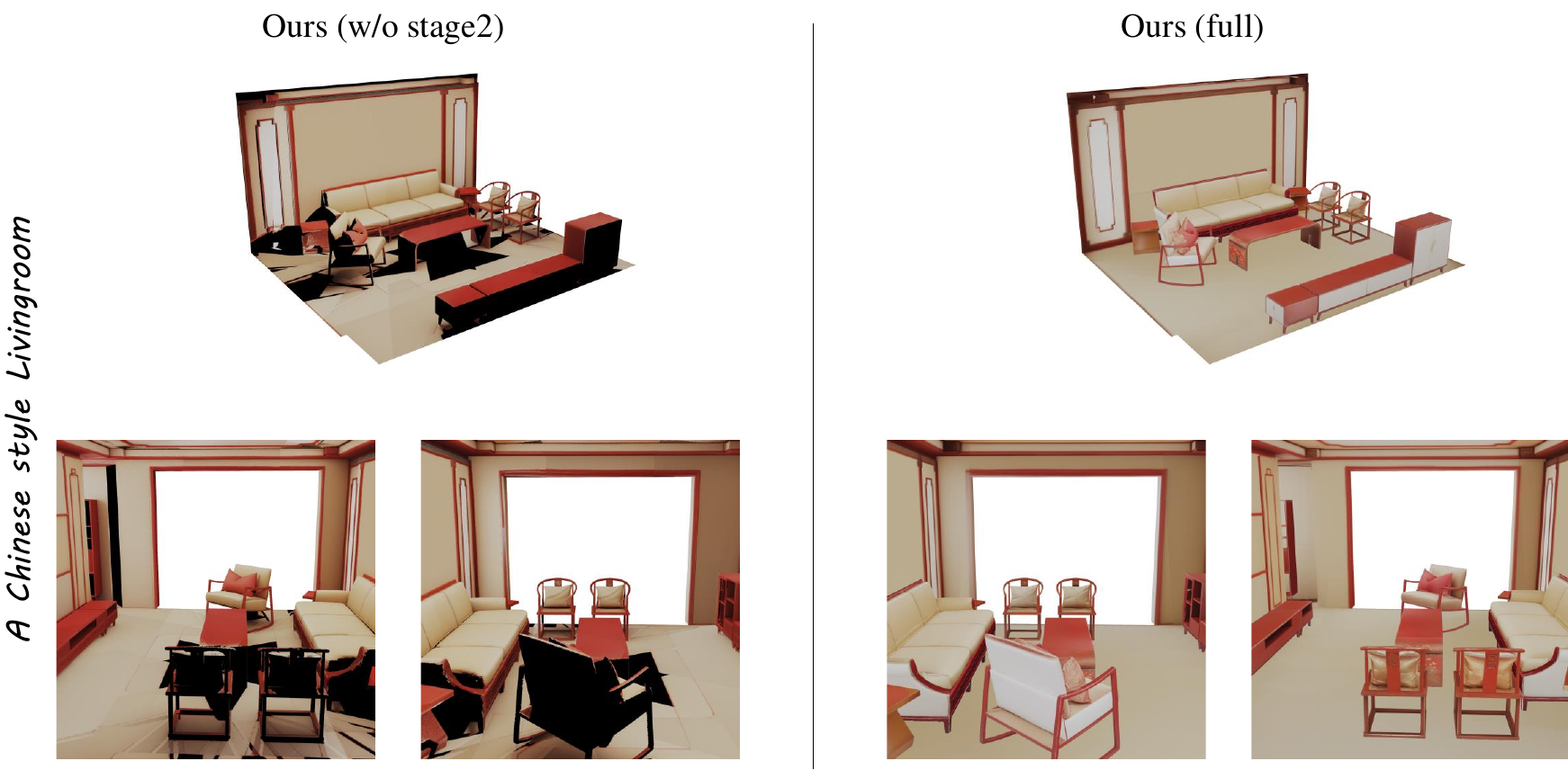}
    \caption{ 
    \textbf{Ablation studies on the stage2.}
         Using only Stage 1 results in large black areas in the texture caused by occlusion, where texture cannot be generated. By incorporating Stage 2, the occlusion issue is resolved while preserving the global style consistency established in Stage 1, and the overall texture quality is significantly improved.
    }
    \label{fig:ablation_stage2}
\end{figure*}

As shown in the Fig.\ref{fig:ablation_stage2}, using only Stage 1 results in large black areas in the texture caused by occlusion, where textures cannot be generated. By incorporating the second stage , the occlusion issue is resolved while preserving the global style consistency established in the first stage, and the overall texture quality is significantly improved.

\section{User Study Details}
\begin{figure*}[!t]
    \centering
    \includegraphics[width=1.0\textwidth]{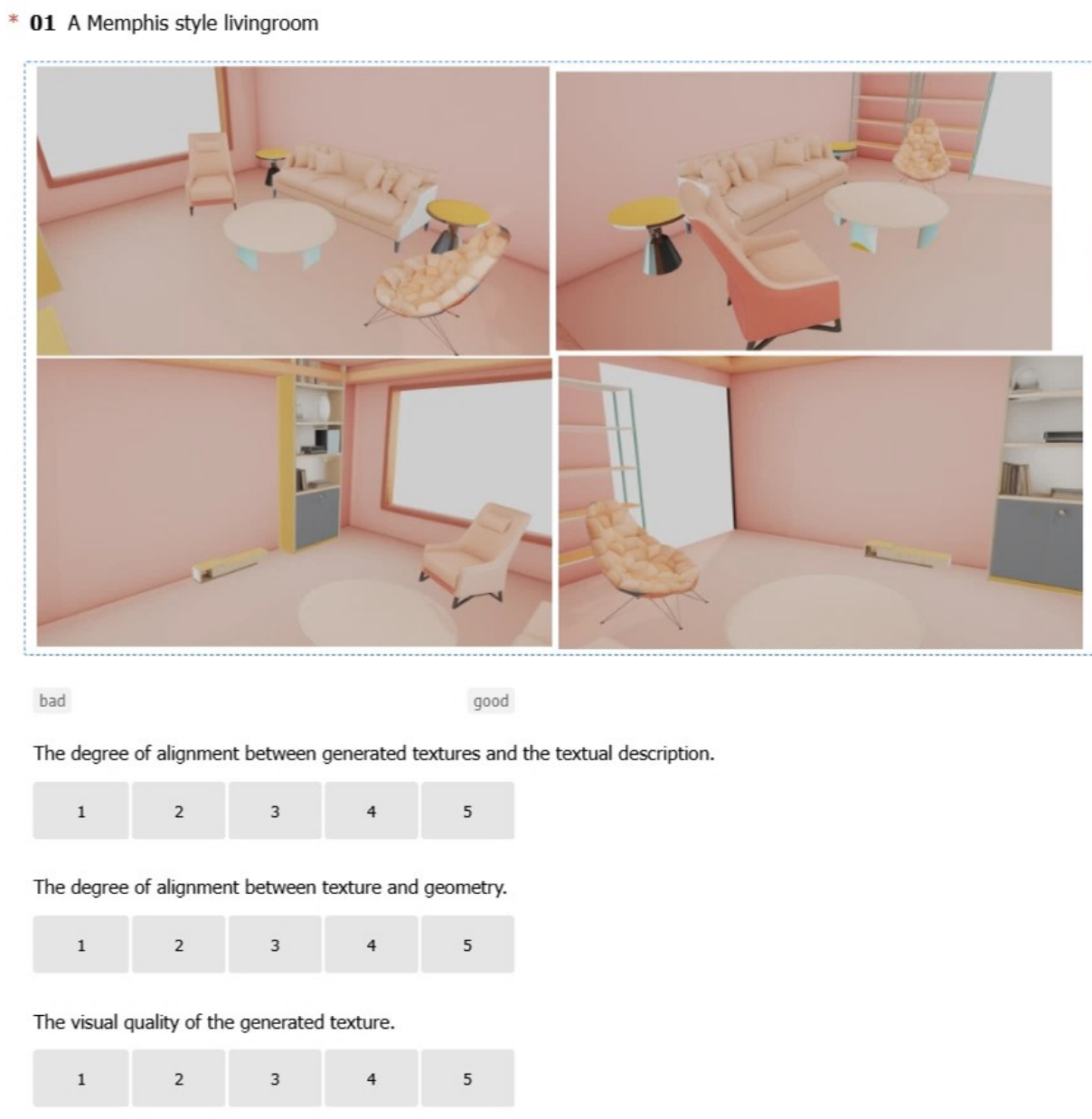}
    \caption{\textbf{The interface of the questionnaire system used in user study.} We present 4 rendered views from 6 different texturing results to each participant and ask them to rate the scenes across three dimensions.}
    \label{fig:interface}
\end{figure*}

We conducted the user study using a web-based questionnaire system to compare our method with three baselines from the human perspective. Fig.\ref{fig:interface} illustrates the interface of our questionnaire system. From the generated results of our method and the baselines, we randomly selected 6 textured scenes created for the same indoor scene using the same textual prompt.
In the interface, we first present the textual prompt used for texture generation, followed by rendered images of the textured scene from four different viewpoints. Participants were then asked to evaluate the generated textures across three dimensions: Prompt Fidelity(PF), Texture and Geometry Alignment(TG) and Visual Quality(VQ), with scores ranging from $1$ (low) to $5$ (high).
In total, we collected 450 ratings from 25 participants and computed the average score for each method. The result is presented in Tab.\ref{tab:user_study}

\section{More visual result} 
\subsection{Different style for same room}
\begin{figure*}[!t]
    \centering
    \includegraphics[width=\linewidth]{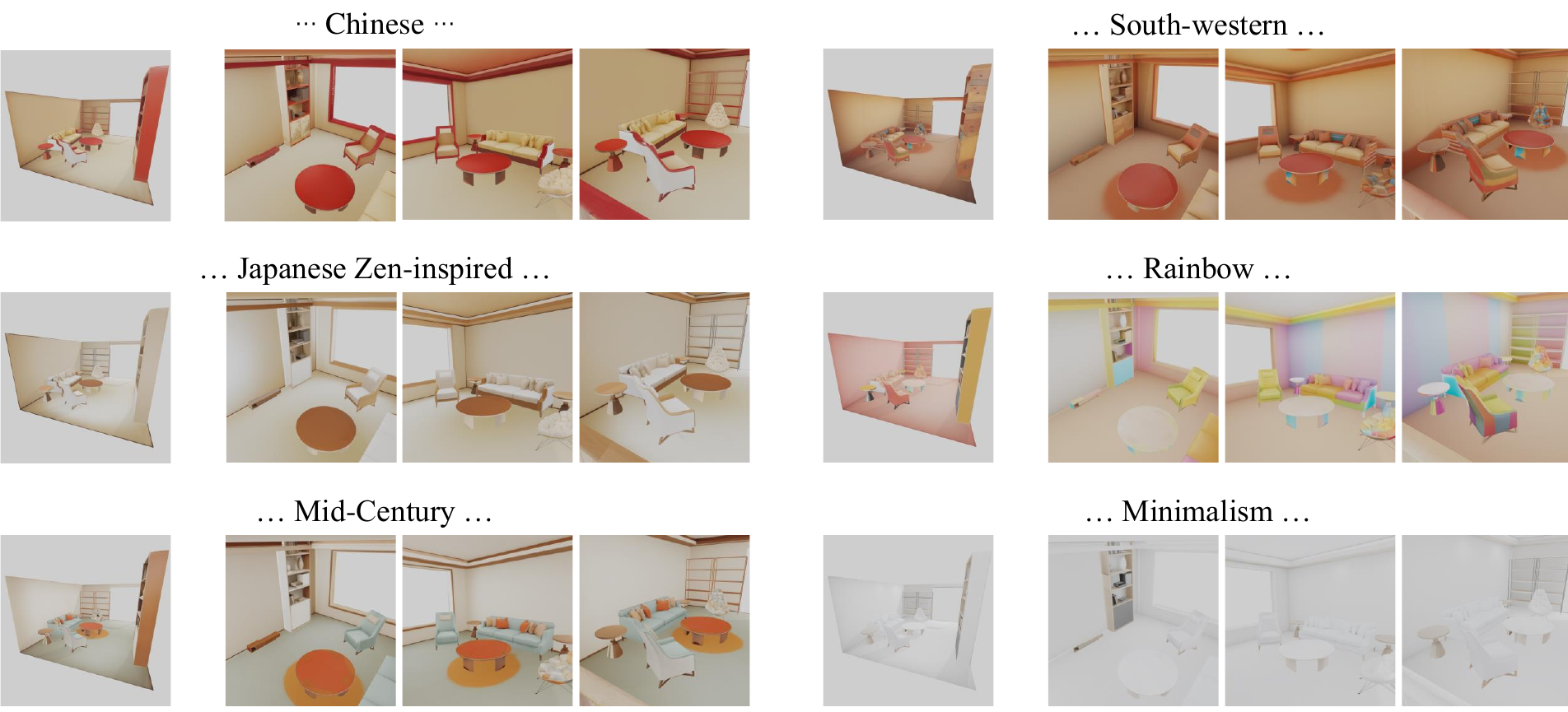}
    \caption{ 
    \textbf{Synthesized texture for 3D-FRONT scenes}
        Our method generates various texture for the same input scene by utilizing the prompt template: 'a 〈STYLE〉 living room' with 6 distinct styles for texture creation.
    }
    \label{fig:same-scene_diff-style}
\end{figure*}

We present more indoor scene texturing results of our method for one room from 3D-Front in Fig.~\ref{fig:same-scene_diff-style}.

\subsection{Visual result on ScanNet++ dataset}
\begin{figure*}[!b]
  \centering
  \includegraphics[width=\linewidth]{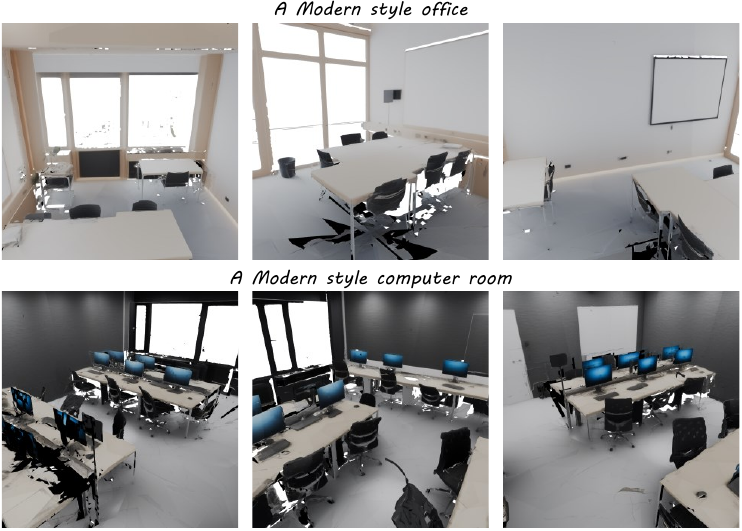}
  \caption{ 
    \textbf{Synthesized texture for ScanNet++ scenes}
        Our method generates similar style texture for different input scene by utilizing the prompt template: `a modern style 〈TYPE〉' with 2 distinct room types for texture creation.
    }
    \label{fig:scannet++}
\end{figure*}

We are providing multi-view visual results of texturing scenes that from ScanNet++ dataset\cite{yeshwanth2023scannet++}. It is worth noting that ScanNet++ features more complex layouts compared to 3D-FRONT, as it is a real-world scan dataset. Visual results are shown in Fig.\ref{fig:scannet++}

\subsection{Example of the intermediate denoising process}
\begin{figure*}[!b]
  \centering
  \includegraphics[width=\linewidth]{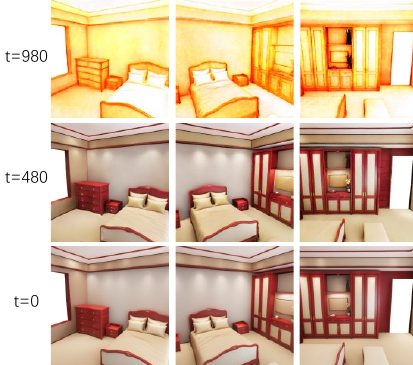}
  \caption{ 
    \textbf{Example of MVIS intermedia result}
         An example of the intermediate of Multi-View Integrated Sampling  process, showcasing the images $x_{0,t}$ across different viewpoints at different steps.
    }
  \label{fig:intermedia_x0}
\end{figure*}

During the MVIS process of generating multiple consistent images, intermediate results emerge from the multi-view sampling process, as shown in the Fig.\ref{fig:intermedia_x0}.
During MVIS process, the $x_{0,t}$ images across different viewpoints at different steps are shown in Fig.\ref{fig:intermedia_x0}

\begin{algorithm*}[!t]
\caption{Multi-view Integrated Repaint Sampling of One Instance}\label{alg:MVRS}
\begin{algorithmic}
\STATE Input: 
    Mesh $\mathcal{M}$, 
    Text $y$, 
    Cameras $\{C^1, \dots, C^N \}$, 
    Textured mask ${\{P^1, \dots, P^N \}}$,
    Stage1 generated texture $\mathcal{T}_{MVIS}$
\STATE Parameters: 
    DDPM noise schedule $\{\sigma_t\}_{t=T}^0$
\vspace{5pt}
\STATE Initialization:
\STATE $\mathcal{T} = 0$
\STATE $\{x_T^{n}\}_{n=1}^{N}\sim \{\mathcal{N}({0}, {I})\}$ \textcolor{blue}{\# Unpainted noisy latents}
\STATE $\{\mathcal{I}_{MVIS}^n \leftarrow \mathcal{R}(\mathcal{T}_{MVIS}, \mathcal{M}, c^n)\}_{n=1}^{N} $ \textcolor{blue}{\# Painted area is colorful,  unpainted area is black}
\STATE $\{x_{MVIS}^{n} \leftarrow \mathcal{E}( \mathcal{I}_{MVIS}^n )\}_{n=1}^{N}$ \textcolor{blue}{\# Painted noise-free latents}
\vspace{5pt}
    \STATE \COMMENT \textbf{Early steps (conditioning on painted area):} 
    \FOR{$t \in \{T \dots T_{end1}\}$}
        \FOR{$n \in \{1 \dots N\}$}
            \STATE $ \epsilon_n \sim \mathcal{N}(\mathrm{0}, \boldsymbol{I})$
            \STATE $ \epsilon_t^n \leftarrow \epsilon_\theta(x_t^n, y, d_n, t)$
            \STATE $ x_{0, t}^n \leftarrow \dfrac{x_t^n - \sqrt{1 - \bar\alpha_t}{\epsilon}_t^n}{\sqrt{\alpha_t}}$
            \STATE $ \mu_{t-1}^n \leftarrow \frac{ \sqrt{\bar\alpha_{t-1}} \beta_t }{ 1-\bar\alpha_t } x_{0, t}^n + \frac{\sqrt{\alpha_t}(1- \bar\alpha_{t-1})}{1-\bar\alpha_i}x_t^n$
            \STATE $ x_{t-1}^n \leftarrow \mu_{t-1}^n + \sigma_t\epsilon_n$ \textcolor{blue}{\# Unpainted latent with t-1 level noise }
            \STATE $ \tilde x_{t-1}^n \leftarrow \sqrt{\bar\alpha_t} x_{MVIS}^{n} + \sqrt{1-\bar\alpha_t}\epsilon_n$ \textcolor{blue}{\# Painted latent with t-1 level noise }
            \STATE $ x_{t-1}^n \leftarrow \tilde x_{t-1}^n \odot P^n + x_{t-1}^n \odot ( 1-P^n )$  \textcolor{blue}{\# Mask combine }
        \ENDFOR
    \ENDFOR
    \vspace{5pt}
    \STATE \COMMENT \textbf{Middle steps (conditioning on painted area):} 
    \FOR{$t \in \{ T_{end1} \dots T_{end2} \}$}
        \FOR{$n \in \{1 \dots N\}$}
            \STATE $ \epsilon_n \sim \mathcal{N}(\mathrm{0}, \boldsymbol{I})$
            \STATE $ \epsilon_t^n \leftarrow \epsilon_\theta(x_t^n, y, d_n, t)$
            \STATE $ x_{0, t}^n \leftarrow \dfrac{x_t^n - \sqrt{1 - \bar\alpha_t}\bar{\epsilon}_t^n}{\sqrt{\alpha_t}}$ \textcolor{blue}{\# Predicted noise-free image at timestep t }
            \STATE $ \mathcal{I}_t^n \leftarrow \mathcal{D}(x_{0, t}^n) $ 
            \STATE $ \mathcal{T}^n \leftarrow \mathcal{R}^{-1}(\mathcal{I}_t^n, \mathcal{M}, C^n) $ \textcolor{blue}{\# Project view-specific noise-free images to view-specific texture maps }
        \ENDFOR\vspace{2pt}
        \STATE $\mathcal{T} = dynamic\_merge (\{ \mathcal{T}^n \}_{n=1}^{N}) $\vspace{2pt} \textcolor{blue}{\# Merge view-specific texture maps to a global-consistent texture }
        \FOR{$n \in \{1 \dots N\}$}
            \STATE $ \tilde x_{0, t}^n \leftarrow \mathcal{E}(\mathcal{R}(\mathcal{T}, \mathcal{M}, C^n))$ \textcolor{blue}{\# Global-consistent noise-free unpainted area latent at timestep t}
            \STATE $ \mu_{t-1}^n \leftarrow \frac{\sqrt{\bar\alpha_{t-1}}\beta_t}{1-\bar\alpha_t} \tilde x_{0, t}^n + \frac{\sqrt{\alpha_t}(1- \bar\alpha_{t-1})}{1-\bar\alpha_t}x_t^n$
            \STATE $ x_{t-1}^n \leftarrow \mu_{t-1}^n + \sigma_t\epsilon_n$ \textcolor{blue}{\# Unpainted latent with t-1 level noise }
            \STATE $ \tilde x_{t-1}^n \leftarrow \sqrt{\bar\alpha_i}x_{MVIS}^{n} + \sqrt{1-\bar\alpha_t}\epsilon_n$ \textcolor{blue}{\# Painted latent with t-1 level noise }
            \STATE $ x_{t-1}^n \leftarrow \tilde x_{t-1}^n \odot P^n + x_{t-1}^n \odot ( 1-P^n )$ \textcolor{blue}{\# Mask combine }
        \ENDFOR
    \ENDFOR
    \vspace{5pt}
    \STATE \COMMENT \textbf{Latter steps (not conditioning on painted area):} 
    \FOR{$t \in \{ T_{end2} \dots 0 \}$}
        \STATE $\mathcal{T} \leftarrow \textbf{MVIS}( \mathcal{M}, y, { \{c^n\}_{n=1}^{N} }, { \{x_t^n\}_{n=1}^{N} })$    
    \ENDFOR
    \STATE $\mathcal{T}_{MVRS} \leftarrow \mathcal{T}$
    
\RETURN Texture map $\mathcal{T}_{MVRS}$
\end{algorithmic}
\end{algorithm*}

\end{document}